# Hybrid Tractable Classes of Binary Quantified Constraint Satisfaction Problems

**Jian Gao, Minghao Yin, Junping Zhou**

College of Computer Science, Northeast Normal University, Changchun, 130024, China.

**Abstract**

In this paper, we investigate the hybrid tractability of binary Quantified Constraint Satisfaction Problems (QCSPs). First, a basic tractable class of binary QCSPs is identified by using the broken-triangle property. In this class, the variable ordering for the broken-triangle property must be same as that in the prefix of the QCSP. Second, we break this restriction to allow that existentially quantified variables can be shifted within or out of their blocks, and thus identify some novel tractable classes by introducing the broken-angle property. Finally, we identify a more generalized tractable class, i.e., the min-of-max extendable class for QCSPs.

## Introduction

As a central problem in artificial intelligence, the Constraint Satisfaction Problem, or CSP in short, has been a popular paradigm to cope with complex combinational problems in recent years. Many kinds of real-world problems can be translated into CSPs. A CSP consists of a set of variables and a set of constraints restricting the combinations of values taken by variables. The structure of each CSP instance is its underlying undirected graph (hyper graph), which is called constraint network. The Quantified Constraint Satisfaction Problem (QCSP) is a generalization of the CSP, where variables are existentially or universally quantified. QCSPs can be employed to represent and solve problems with uncertainty (Gent et al. 2008), such as game playing, dynamic scheduling, and model checking.

It is well-known that deciding the satisfiability of CSPs is NP-complete. In this sense, there exists no polynomial algorithm to solve CSPs unless P=NP. Hence, a great amount of attention has been drawn on identifying polynomial-time tractable subclasses of CSPs. There are three main kinds of tractable classes: structural classes, where constraint networks have some special structures; relational classes, where only forms of constraints are restricted; hybrid classes, which depend on both the structures of the networks and the forms of constraints (Dechter 2003). Furthermore, QCSPs have been proved to be PSPACE-complete (Börner et al. 2003). In the past decade, we have also viewed great effort made to identify tractable classes of QCSPs. Obviously, the analogous class of QCSPs cannot be tractable if a class of CSPs does not ensure the tractability (Chen 2004b). Therefore, it is not strange that most of the literatures focus on deciding whether a quantified extension of an existing tractable CSP class remains tractable. Early works along this line are quantified 2-satisfiability (Aspvall, Plass, and Tarjan 1979) and quantified horn satisfiability (Karpinski, Büning, and Schmitt 1987), which are proved to be tractable. Recently, Börner et al. (2003) and Chen (2004a) have provided some tractable classes by exploiting particular relations. Chen (2004b) has also presented structural tractable classes having bounded treewidth. Gottlob, Greco, and Scarcello (2005) have introduced a clear picture of the tractability/intractability frontier for structural classes of QCSPs, and identified new structural classes by considering quantifiers as part of the scope structure. From those results, we can learn that though those classes of QCSPs have their prototypes on CSPs, some traditional methods for identifying tractable classes of CSPs may not help a lot, and thus more restrictions are usually required to ensure the tractability of QCSPs.

Though an increasing number of relational and structural tractable classes of QCSPs have been identified, there is little attention on hybrid tractable classes, which have achieved great success on CSPs (van Beek and Dechter 1995; Cohen et al. 2011; Zhang and Freuder 2008; Cooper, Jeavons, and Salamon 2010; Cooper and Zivny 2010). We say a subclass of QCSPs is tractable if the satisfiability of all its instances can be determined in polynomial time and if identifying whether those instances belong to the subclass can achieve in polynomial time. Thus, some interesting question arise: can we identify some tractable classes of QCSPs restricted by hybrid techniques and what extra restrictions are required? Can we expand the tractability frontier by generalizing these hybrid techniques so that more novel tractable classes can be identified? The aim of this paper is to answer those questions. We first identity a basic tractable class of QCSPs with the Broken-Triangle Property (BTP) (Cooper, Jeavons, and Salamon 2010), where the variable ordering of broken-triangle property is same as the variable ordering in the prefix. Second, we investigate the class allowing that the variable ordering can differ from the ordering in the prefix, and propose the broken-angle property for QCSPs so as to identify a novel tractable class. At last, we extend those properties to class with the min-of-max extensibility and identify more hybrid tractable classes in a similar way.

# Preliminaries

## Problem Definitions

We first give a formal definition of the QCSPs.

**Definition 1.** A QCSP $P$ is a tuple $<V, Q, \Pi, D, C>$ where:

$V$ is a set of $n$ variables.

$Q$ is a mapping from $V$ to the set of quantifiers $\{\exists, \forall\}$. For each variable $x_i \in V$, $Q(x_i)$ is a quantifier ($\exists$ or $\forall$) associated with $x_i$.

$\Pi$ is a one-to-one mapping from $V$ to the ordered set $\{1, ..., n\}$. For each variable $x_i \in V$, $\Pi(x_i)$ is a number in the ordered set, and we denote $\Pi^{-1}$ the inverse mapping of $\Pi$, where $\Pi^{-1}(i)$ is a variable in $V$.

$D$ is a mapping from $V$ to a set of domains $D = \{D(x_1), ..., D(x_n)\}$. For each variable $x_i \in V$, $D(x_i)$ is the finite domain of its possible values.

$C = \{C_1, ..., C_e\}$ is a set of $t$ constraints. Each constraint $C_i \in C$ is defined as a pair $(vars(C_i), rel(C_i))$, where $vars(C_i) = (x_{j_1}, ..., x_{j_k})$ is an ordered subset of $V$ called the constraint scope. The size of $vars(C_i)$ is called the arity of $C_i$; $rel(C_i)$ is a subset of the *Cartesian* product $D(x_{j_1}) \times ... \times D(x_{j_k})$ and it specifies the allowed combinations of values for the variables in $vars(C_i)$. For the simplicity, we always use the notion $R_i$ to denote $rel(C_i)$ for short in the rest of the paper.

As we need to investigate relations between problems with different variable ordering, $\Pi$ is introduced, so our definition employs 5-tuple rather than 4-tuple in the literature (Gent et al. 2008). A QCSP $P = (V, Q, \Pi, D, C)$ represents the logical formula $\phi = Q(x_{\Pi^{-1}(1)})x_{\Pi^{-1}(1)}...Q(x_{\Pi^{-1}(n)})x_{\Pi^{-1}(n)}(C)$, where $Q(x_{\Pi^{-1}(1)})x_{\Pi^{-1}(1)}...Q(x_{\Pi^{-1}(n)})x_{\Pi^{-1}(n)}$ is the prefix of $P$.

Given a QCSP $P = (V, Q, \Pi, D, C)$, let $d$ be the maximum size of domains; $P$ is binary if for each constraint $C_i \in C$, $|vars(C_i)| = 2$. Especially, we use $C_{ij}$ to denote the binary constraint between variables $x_i$ and $x_j$ in $P$. We say $\beta \in D(x_j)$ supports $\alpha \in D(x_i)$ if $(\alpha, \beta) \in R_{ij}$, and $R_{ij}(\alpha) = \{\beta | \beta$ supports $\alpha$, and $\beta \in D(x_j)\}$. The assignment of value $\alpha \in D(x_i)$ to variable $x_i$ is denoted by $<x_i, \alpha>$. A *strategy* is a tree with each level of the tree corresponding to a variable in $P$. The level $i$ corresponds to the variable $x_{\Pi^{-1}(i)}$ in $V$, while level 0 (the root) is associated with an implicit variable $x_0$. Each node in level 1 to $n$ is assigned by a value from the domain of corresponding variable. The number of children of a node in level $i$ ($i < n$) is equal to the size of $D(x_{\Pi^{-1}(i+1)})$ if $Q(x_{\Pi^{-1}(i+1)}) = \forall$; otherwise, the number is 1. A *scenario* is a sequence of assignments to all variables (all nodes on a path from the root to a leaf node in a strategy). Thus, there are as many scenarios of a strategy as leaves of the strategy. A scenario is consistent if the assignments satisfy all constraints in the QCSP, and a strategy is consistent if all the scenarios in the strategy are consistent. A consistent strategy is also called a *solution* to the QCSP. A *subtree* for a node $w$ is defined as a connected subgraph consisting of all nodes on the path from the root to $w$ and all of its descendant nodes. A *partial strategy* is a tree that includes level 0 and the next $k$ levels of a strategy, where level $i$ ($1 \leq i \leq k$) corresponds to the variable $x_{\Pi^{-1}(i)}$. A *partial scenario* of a partial strategy including $k$ variables is a sequence of assignments to those $k$ variables, and it the partial scenario (or partial assignment) is consistent if every constraint covered by it is satisfied.

Here we define some variable sets as follows: $block(\Pi, x_i)$ is a maximal variable set such that for each $x_j$ in it, $Q(x_i) = Q(x_j)$ and there does not exist a variable $x_k$ such that $x_k$ is between $x_i$ and $x_j$ with respect to $\Pi$ and $Q(x_i) \neq Q(x_k)$ ($x_j$ and $x_i$ may be the same variable); $pre_\forall(\Pi, x_i)$ ($pre_\exists(\Pi, x_i)$) is a set consisting of all the universally (existentially) quantified variables before $x_i$ with respect to $\Pi$ but not in $block(\Pi, x_i)$; $suc_\forall(\Pi, x_i)$ ($suc_\exists(\Pi, x_i)$) is a set consisting of all the universally (existentially) quantified variables after $x_i$ with respect to $\Pi$ but not in $block(\Pi, x_i)$; $suc(\Pi, x_i)$ is a set consisting of all variables after $x_i$ with respect to $\Pi$; $x_{\forall i}^{\Pi}$ is the variable such that if $pre_\forall(\Pi, x_i)$ is not empty, $x_{\forall i}^{\Pi}$ is the closet universally quantified variable before $x_i$ with respect to $\Pi$, namely, $x_{\forall i}^{\Pi} \in pre_\forall(\Pi, x_i)$ and for each $x_j \in pre_\forall(\Pi, x_i)$ ($x_j \neq x_{\forall i}^{\Pi}$), $\Pi(x_j) < \Pi(x_{\forall i}^{\Pi})$, otherwise $x_{\forall i}^{\Pi}$ is $x_0$.

## Consistency in QCSPs

Quantified consistency for QCSPs was investigated for solving QCSPs. Gent et al. (2008) proposed the *Quantified Arc Consistency* (QAC), and Stergiou (2008) proposed the *Directional Quantified Arc Consistency* (DQAC) and *directional quantified path consistency*. Following these definitions, we give the definition of quantified arc consistency and directional quantified $k$-consistency:

**Definition 2.** The binary QCSP $P = <V, Q, \Pi, D, C>$ is quantified arc consistent iff for each pair of variables $(x_i, x_j)$, where $\Pi(x_i) < \Pi(x_j)$, each $C_{ij} \in C$ must satisfy one of the following cases:

$\exists x_i \exists x_j$ For each $\alpha \in D(x_i)$, there exists at least one value $\beta \in D(x_j)$ such that $(\alpha, \beta) \in R_{ij}$; for each $\beta \in D(x_j)$, there exists at least one value $\alpha \in D(x_i)$ such that $(\alpha, \beta) \in R_{ij}$.

$\forall x_i \forall x_j$ For each $\alpha \in D(x_i)$, any value $\beta \in D(x_j)$ must satisfy $(\alpha, \beta) \in R_{ij}$; for each $\beta \in D(x_j)$, any value $\alpha \in D(x_i)$ must satisfy $(\alpha, \beta) \in R_{ij}$.

$\forall x_i \exists x_j$ For each $\alpha \in D(x_i)$, there exists at least one value $\beta \in D(x_j)$ such that $(\alpha, \beta) \in R_{ij}$; for each $\beta \in D(x_j)$, there exists at least one value $\alpha \in D(x_i)$ such that $(\alpha, \beta) \in R_{ij}$.

$\exists x_i \forall x_j$ For each $\alpha \in D(x_i)$, any value $\beta \in D(x_j)$ must satisfy $(\alpha, \beta) \in R_{ij}$; for each $\beta \in D(x_j)$, there exists at least one value $\alpha \in D(x_i)$ such that $(\alpha, \beta) \in R_{ij}$.

A pair $(x_i, x_j)$ of variables ($\Pi(x_i) < \Pi(x_j)$) is quantified arc consistent if it satisfies one of the above cases.

**Definition 3.** *The binary QCSP $P = (V, Q, \Pi, D, C)$ is directional quantified k-consistent iff for every k-tuple $(x_{m_1}, ..., x_{m_{k-1}}, x_{m_k})$ such that $\Pi(x_{m_i}) < \Pi(x_{m_j})$ ($1 \leq i < j \leq k$), one of the following cases must be satisfied:*

$\lambda x_{m_1}...\lambda x_{m_{k-1}} \exists x_{m_k}$: *For each consistent partial assignment $(\alpha_1, ..., \alpha_{k-1})$, there exists at least one value $\gamma \in D(x_{m_k})$ such that $(\alpha_i, \gamma) \in R_{m_i m_k}$ ($1 \leq i \leq k-1$).*

$\lambda x_{m_1}...\lambda x_{m_{k-1}} \forall x_{m_k}$: *For each consistent partial assignment $(\alpha_1, ..., \alpha_{k-1})$, each $\gamma \in D(x_{m_k})$ must satisfy $(\alpha_i, \gamma) \in R_{m_i m_k}$ ($1 \leq i \leq k-1$).*

*where $\lambda$ represents either quantifier in $\{\exists, \forall\}$.*

In particular, directional quantified 2-consistency is usually called directional quantified arc consistency. A binary QCSP is *strong, directional quantified k-consistent* if it is directional quantified $i$-consistent for all $i \leq k$; a binary QCSP is *directional quantified globally consistent* if it is strong, directional quantified n-consistent.

Gent et al. (2008) have presented a quantified arc consistency algorithm called QAC-2001. The time complexity of the algorithm is $O(d^2 e)$. An algorithm for strong directional quantified path consistency has been proposed by Stergiou (2008), whose time complexity is $O(n^3 d^3)$ in the worst case. We regard that a QCSP is empty if a variable domain is empty after applying consistency algorithm to the problem.

If $P = (V, Q, \Pi, D, C)$ is directional quantified globally consistent, a solution to $P$ can be constructed as follows. Create a node as the root in level 0. Then, for each level $i$, if $Q(x_{\Pi^{-1}(i)}) = \exists$, for each node $w$ in level $i$-1, create a node as the child of $w$ and label it with a consistent value from $D(x_{\Pi^{-1}(i)})$; if $Q(x_{\Pi^{-1}(i)}) = \forall$, for each node $w$ in level $i$-1, create $|D(x_{\Pi^{-1}(i)})|$ nodes as the children of $w$ and label them with each value in $D(x_{\Pi^{-1}(i)})$ respectively. After all levels are extended, the tree constructed is a solution to $P$.

## Basic Tractable Class

In this section, we will discuss the basic tractable properties based on the BTP, which is first introduced by Cooper, Jeavons, and Salamon (2008). For details of the BTP, we refer to (Cooper, Jeavons, and Salamon 2008) and (Cooper, Jeavons, and Salamon 2010). In this paper, in order to propose a basic tractable class of QCSPs, we first define the BTP for QCSPs.

**Definition 4.** *A binary QCSP $P = (V, Q, \Pi, D, C)$ satisfies the broken-triangle property (QBTP for short), if, for all triples of variables $(x_i, x_j, x_k)$ such that $\Pi(x_i) < \Pi(x_j) < \Pi(x_k)$, if $(\alpha, \beta) \in R_{ij}$, $(\alpha, \gamma) \in R_{ik}$ and $(\beta, \theta) \in R_{jk}$, then either $(\alpha, \theta) \in R_{ik}$ or $(\beta, \gamma) \in R_{jk}$.*

The definition of the QBTP is similar to that of the BTP for classical CSP, but it restricts that the variable ordering for the QBTP must be same as that of the prefix in the QCSP.

**Lemma 1.** *Let $P = (V, Q, \Pi, D, C)$ be a binary QCSP. If $P$ satisfies the QBTP, for all triples of variables $(x_i, x_j, x_k)$ such that $\Pi(x_i) < \Pi(x_j) < \Pi(x_k)$, then for all $(\alpha, \beta) \in R_{ij}$ ($R_{ik}(\alpha) \subseteq R_{jk}(\beta)$) or ($R_{jk}(\beta) \subseteq R_{ik}(\alpha)$).*

The proof of Lemma 1 is similar with that of Lemma 2.4 in (Cooper, Jeavons, and Salamon 2010), therefore we omit the proof.

**Lemma 2.** *Let $P = (V, Q, \Pi, D, C)$ be a binary QCSP that satisfies the QBTP, if the reduced problem $P'$ after enforcing quantified arc consistency on $P$ is not empty, $P'$ satisfies the QBTP.*

Lemma 2 is also obvious and direct. It can be seen that only domain restrictions (i.e., removing values of a domain) are required in the QAC algorithm. According to the definition of the QBTP, it is easy to see that the QBTP will be preserved after the domain restrictions. In this sense, we can draw the conclusion that $P'$ also satisfies the QBTP if it is not empty.

**Theorem 1.** *Let $P = (V, Q, \Pi, D, C)$ be a binary QCSP. If $P$ satisfies the QBTP, the satisfiability of $P$ can be determined by a polynomial-time algorithm.*

*Proof.* We first enforce QAC on $P$, which can be computed in polynomial time. If the result is empty, $P$ is unsatisfiable. Otherwise, we denote the reduced equivalent problem by $P' = (V, Q, \Pi, D', C')$. According to Lemma 2, $P'$ satisfies the QBTP. Next, we should show that $P'$ is directional quantified globally consistent. Suppose $P'$ is directional quantified ($k$-1)-consistent ($k \leq n$). Without loss of generality, let $(\alpha_1, ..., \alpha_{k-1})$ be consistent partial assignments to $(x_{m_1}, ..., x_{m_{k-1}})$ and $x_{m_k}$ be any variable such that $\Pi(x_{m_i}) < \Pi(x_{m_k})(1 \leq i \leq k-1)$. If $Q(x_{m_k}) = \forall$, as $P'$ is quantified arc consistent, all values in $D'(x_{m_k})$ support all values in $D'(x_{m_i})$, where $1 \leq i \leq k-1$. Thus, all values in $D'(x_{m_k})$ support the assignments $(\alpha_1, ..., \alpha_{k-1})$. If $Q(x_{m_k}) = \exists$, $\bigcap_{1 \leq i \leq k-1} C'_{m_i m_k}(\alpha_i)$ is not empty because of the QBTP and Lemma 1, so there exists at least one value supporting the assignments $(\alpha_1, ..., \alpha_{k-1})$ to $(x_{m_1}, ..., x_{m_{k-1}})$. Hence, $P'$ is directional quantified $k$-consistent. As $k$ can take an arbitrary integer no more than $n$, and $P'$ is directional quantified 2-consistent, it is directional quantified globally consistent. Therefore, $P'$ is satisfiable, so is $P$. To sum up, the satisfiability of $P$ can be determined by QAC-2001, which is a polynomial-time algorithm. □

**Proposition 1.** *Let $P = (V, Q, \Pi, D, C)$ be a binary QCSP, there exists a polynomial-time algorithm to determine whether $P$ satisfies the QBTP.*

Proposition 1 is obvious, because checking whether each triple of ordered variables satisfies the QBTP requires $O(d^4)$ time. Therefore, given a binary QCSP, determining whether it satisfies the QBTP can be done by an algorithm in $O(d^4 n^3)$. If it satisfies the QBTP, the satisfiability of such problem can be specified in $O(d^2 t)$ time at worst.

# Extended Tractable Classes

Definition 4 restricts the variable ordering for the QBTP identical to the prefix ordering. In this section, we show that a QCSP may also be tractable without such strong restriction.

**Block-compatible orderings**

First, we consider the case that variables shift within blocks.

**Definition 5.** *Let $P^\Pi = (V, Q, \Pi, D, C)$ and $P^\Delta = (V, Q, \Delta, D, C)$ be binary QCSPs. $\Pi$ is block-compatible with $\Delta$ if they satisfy the following conditions*:
  1. *For each existentially quantified variable $x_i \in V$, $pre_\forall(\Pi, x_i) = pre_\forall(\Delta, x_i)$;*
  2. *For each pair of universally quantified variables $(x_i, x_j)$, if $\Pi(x_i) < \Pi(x_j)$, then $\Delta(x_i) < \Delta(x_j)$.*

As can be seen from Definition 5, we only consider the ordering of existentially quantified variables. The reason is that constraints between two universally quantified variables make no effect on the QBTP. As existentially quantified variables only shift in their own blocks, we have $block(\Pi, x_i) = block(\Delta, x_i)$ for each $x_i \in V$. Based on Definition 5, we discuss the resulting tractable class as follows.

**Theorem 2.** *Given a binary QCSP $P^\Pi = (V, Q, \Pi, D, C)$, the satisfiability of $P^\Pi$ can be determined by a polynomial-time algorithm if there exists a mapping $\Delta$ such that*:
  1. *$\Pi$ is block-compatible with $\Delta$;*
  2. *$P^\Delta = (V, Q, \Delta, D, C)$ satisfies the QBTP.*

*Proof.* Clearly, if there exists a pair $(x_i, x_j)$ that is not quantified arc consistent in $P^\Delta$, it is not quantified arc consistent in $P^\Pi$, so $P^\Pi$ is unsatisfiable; otherwise, since $P^\Delta$ is not empty after enforcing QAC and it satisfies the QBTP, $P^\Delta$ has a solution, denoted by $s$, according to Theorem 1. Because $\Pi$ is block-compatible with $\Delta$, we have $block(\Pi, x_i) = block(\Delta, x_i)$. So we can construct a solution $s'$ to $P^\Pi$ from $s$ by reordering each level in $s$ such that the new variable ordering in $s'$ is same as that in the prefix of $P^\Pi$. Because variables are only shifted in their own blocks, nodes in a path of $s$ can be relinked in the new ordering to construct a path of $s'$. In this way, we can obtain $s'$ by relinking each path. Considering any scenario in $s'$, it is consistent because we can find a scenario in $s$ with same assignments to all variables. So $s'$ is a solution to $P^\Pi$. Therefore, we can conclude that the satisfiability of $P^\Pi$ can be determined in polynomial time by enforcing QAC on $P^\Delta$. □

**Semi-compatible orderings**

Next, we discuss the case that variables shift between blocks. As we have discussed in the former subsection, universally quantified variables are also restricted by the original orderings. Therefore, we should consider two situations: shifting existentially quantified variables in front of universally quantified variables, or shifting existentially quantified variables after universally quantified variables. The former is trivial, because the existentially quantified variables moved forward must satisfy quantified arc consistency. For example, suppose that $\forall x_i \exists x_j C_{ij}$ is a subproblem. When $x_j$ moves forward we have a subproblem $\exists x_j \forall x_i C_{ji}$. We hope that each value in $D(x_j)$ should be supported by all values in $D(x_i)$, i.e. it is quantified arc consistent, which is too rigorous to satisfy. Besides, it is evident that $\forall x_i \exists x_j C_{ij}$ will satisfy the QBTP if $\exists x_j \forall x_i C_{ji}$ is quantified arc consistent. So we only need to consider the latter.

**Definition 6.** *Let $P^\Pi = (V, Q, \Pi, D, C)$ and $P^\Delta = (V, Q, \Delta, D, C)$ be binary QCSPs, $\Pi$ is semi-compatible with $\Delta$ if they satisfy the following conditions*:
  1. *For each existentially quantified variable $x_i \in V$, $pre_\forall(\Pi, x_i) \subseteq pre_\forall(\Delta, x_i)$;*
  2. *For each pair of universally quantified variables $(x_i, x_j)$, if $\Pi(x_i) < \Pi(x_j)$, then $\Delta(x_i) < \Delta(x_j)$.*

**Definition 7.** *Let $P^\Pi = (V, Q, \Pi, D, C)$ be a binary QCSP and $s$ be a consistent partial strategy with $k$ variables of $P^\Pi$. Given a universally quantified variable $x_i$ and an existentially quantified variable $x_j$ such that $\Pi(x_i) < \Pi(x_j) \leq k$, $s$ is $(x_i,x_j)$-compatible if for each node $w$ in level $\Pi(x_i)$ of $s$, all nodes in level $\Pi(x_j)$ of $w$'s subtree have the same label.*

Definition 7 restricts $x_i$ is universally quantified and $x_j$ is existentially quantified, as in the rest of this paper, we only care about $(x_i,x_j)$-compatibility, where $Q(x_j) = \exists$ and $x_i$ is the closest universally quantified variable before $x_j$ with respect to a given variable ordering. We give an example to illustrate the Definition 7.

**Example 1.** Given the problem $P^\Pi = (V, \Pi, D, C)$ with 5 variables, where $Q(x_1) = \forall$, $Q(x_2) = \exists$, $Q(x_3) = \forall$, $Q(x_4) = \exists$, $Q(x_5) = \exists$, $\Pi(x_i) = i (1 \leq i \leq 5)$. Suppose The tree in Fig. 1 is a solution (a consistent partial strategy with 5 variables) to it, then there two subtrees of $x_1$, and the solution is $(x_1,x_4)$-compatible while it is not $(x_1,x_5)$-compatible because of the different labels in level $x_5$ of the left subtree.

**Lemma 3.** *Let $P^\Pi = (V, Q, \Pi, D, C)$ and $P^\Delta = (V, Q, \Delta, D, C)$ be binary QCSPs such that $\Pi$ is semi-compatible with $\Delta$. If there exists a solution $s$ to $P^\Delta$ such that for each existentially quantified variable $x_i \in V$, $s$ is $(x^\Pi_{\forall i}, x_i)$-compatible, then $P^\Pi$ is satisfiable.*

*Proof.* We first construct an identical strategy $s'$ of $s$. For each $x_i$ such that $Q(x_i) = \exists$, we shift the level of $x_i$ in $s'$ to the position after the level of $x^\Pi_{\forall i}$ by the following method: as $\Pi$ is semi-compatible with $\Delta$, $\Delta(x^\Pi_{\forall i}) < \Delta(x_i)$. Thus, for each node $w$ in the level of $x_{\forall i}^{\Pi}$ in $s'$, in the subtree of $w$, there should exist some node(s) in the level of $x_i$, and all such node(s) are labeled by a common assignment, denoted by $<x_i, \alpha>$, because of $s$ is $(x_{\forall i}^{\Pi}, x_i)$-compatible. We remove these nodes (or this node) and create a node labeled by $<x_i, \alpha>$ as the child of $w$, then relink all paths in the subtree keeping its scenarios unchanged. After all existentially quantified variables are shifted we get the strategy $s'$ with a new variable ordering denoted by $x_{\Omega^{-1}(1)} x_{\Omega^{-1}(2)} \ldots x_{\Omega^{-1}(n)}$. Clearly, each scenario in $s'$ has a corresponding scenario in $s$ with the same assignments to all variables, so all scenarios in $s'$ are consistent. Hence, $s'$ is a solution to $P^{\Omega} = (V, Q, \Omega, D, C)$.

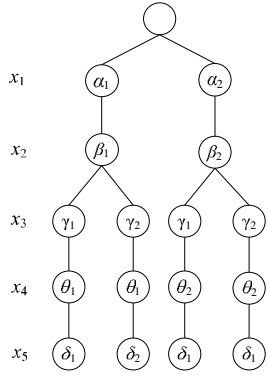

Figure 1: A solution tree of Example 1.

Next, we prove $\Pi$ is block-compatible with $\Omega$. Since $\Pi$ is semi-compatible with $\Delta$ and the ordering of universally quantified variables do not change during the construction of $s'$, for each pair of universally quantified variables $(x_i, x_j)$ such that $\Pi(x_i) < \Pi(x_j)$, we have $\Omega(x_i) < \Omega(x_j)$. Moreover, from the construction of $s'$, we can see that each existentially quantified variable $x_i \in V$ is moved to the block after $x_{\forall i}^{\Pi}$ because no universally quantified variable is between $x_{\forall i}^{\Pi}$ and $x_i$ with respect to $\Omega$, and then we have $pre_\forall(\Pi, x_i) = pre_\forall(\Omega, x_i)$. Therefore, $\Pi$ is block-compatible with $\Omega$. Then, we can construct a solution to $P^\Pi$ from $s'$ according to Theorem 2.

As a result, we can conclude that $P^\Pi$ is satisfiable. □

Lemma 3 provides a method for transforming solution tree when shifting a compatible existential variable to a former block. The method also guarantees the new tree is consistent to the problem after shifting. Consider Example 1, when we shift $x_4$ to the position after $x_1$, a solution to the problem with the prefix $\exists x_1 \forall x_4 \exists x_2 \forall x_3 \exists x_5$ can be constructed by the transforation, which is shown in Fig. 2.

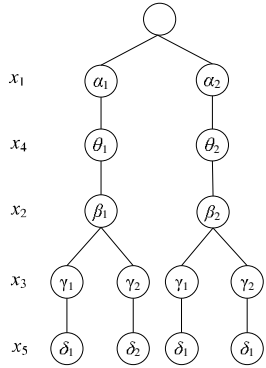

Figure 2: A transformed tree of Example 1.

**Lemma 4.** *Let $P^\Pi = (V, Q, \Pi, D, C)$ and $P^\Delta = (V, Q, \Delta, D, C)$ be binary QCSPs, if $\Pi$ is semi-compatible with $\Delta$ and if $P^\Pi$ is quantified arc consistent, then $P^\Delta$ is quantified arc consistent.*

*Proof.* For any pair of $(x_i, x_j)$, if $\Pi(x_i) < \Pi(x_j)$ and $\Delta(x_i) < \Delta(x_j)$, it is clear that $(x_i, x_j)$ is quantified arc consistent in $P^\Delta$; if $\Pi(x_i) < \Pi(x_j)$ and $\Delta(x_i) > \Delta(x_j)$, since $\Pi$ is semi-compatible with $\Delta$, we only need to consider the following two cases. When $Q(x_i) = \exists$ and $Q(x_j) = \forall$, with the fact that $P^\Pi$ is quantified arc consistent, any value in $D(x_i)$ is supported by all values in $D(x_j)$, so each value in $D(x_j)$ is supported by at least one value in $D(x_i)$, and also each value in $D(x_i)$ is supported by at least one value in $D(x_j)$. When $Q(x_i) = \exists$ and $Q(x_j) = \exists$, it is obvious that the quantified arc consistency holds. Therefore, $P^\Delta$ is quantified arc consistent. □

**Definition 8.** *A triple of variables $(x_i, x_j, x_k)$ satisfies the broken-angle property under a binary QCSP $P^\Pi = (V, Q, \Pi, D, C)$ (QBAP for short) if*:

1. $\Pi(x_i) \leq \Pi(x_j) < \Pi(x_k)$;
2. *For each pair of $(\alpha, \beta)$ ($\alpha \in D(x_i), \beta \in D(x_j)$), if $(\alpha, \gamma) \in R_{ik}$ and $(\beta, \theta) \in R_{jk}$, then either $(\alpha, \theta) \in R_{ik}$ or $(\beta, \gamma) \in R_{jk}$.*

The QBAP is stronger than the QBTP. Any pair of $(\alpha, \beta)$ must satisfy the property in Definition 8. $\alpha$ and $\beta$ may be in a same variable domain, and $(\alpha, \beta)$ does not have to be in $R_{ij}$ if $x_i \neq x_j$. Similar to Lemma 1, we have $R_{ik}(\alpha) \subseteq R_{jk}(\beta)$ or $R_{jk}(\beta) \subseteq R_{ik}(\alpha)$. Based on the QBTP and the QBAP, we propose a new hybrid tractable class, more general than those introduced in Theorem 1 and in Theorem 2.

**Theorem 3.** *Given a binary QCSP $P^\Pi = (V, Q, \Pi, D, C)$ that is quantified arc consistent, $P^\Pi$ is satisfiable if there exists a mapping $\Delta$ such that*:

1. *$\Pi$ is semi-compatible with $\Delta$;*
2. *$P^\Delta = (V, Q, \Delta, D, C)$ satisfies the QBTP;*
3. *For each existentially quantified variable $x_k$, for each pair of variables $(x_i, x_j)$ such that $x_i, x_j \in dif(\Pi, \Delta, x_k)$ and $\Delta(x_i) \leq \Delta(x_j)$, the triple $(x_i, x_j, x_k)$ satisfies the QBAP under $P^\Delta$, where $dif(\Pi, \Delta, x_k) = suc_\exists(\Pi, x_k) - suc(\Delta, x_k)$.*

*Proof.* As $P^\Pi$ is quantified arc consistent and $\Pi$ is semi-compatible with $\Delta$, $P^\Delta$ is quantified arc consistent according to Lemma 4. $P^\Delta$ satisfies the QBTP, so $P^\Delta$ is directional quantified globally consistent, and we can construct a solution to $P^\Delta$.

We then show that there exists a solution $s$ to $P^\Delta$ such that for each existentially quantified variable $x_i \in V$, $s$ is $(x_{\forall i}^\Pi, x_i)$-compatible. To construct such a solution $s$, we should modify some steps of the method for constructing solutions to QCSPs. It is easy to construct a consistent partial strategy with 1 variable. We then suppose there exists a consistent partial strategy with $k$ variables ($1 \leq k < n$) such that for each existentially quantified variable $x_i$ ($\Delta(x_i) \leq k$) the partial strategy is $(x_{\forall i}^\Pi, x_i)$-compatible. It is clear that extending the level $k + 1$ by a universally quantified variable can be achieved by the ordinary way mentioned in the preliminaries. But to extend the level $k + 1$ with an existentially quantified variable (we denote the variable $x_{\Delta^{-1}(k+1)}$ by $x_l$ for short, where $l = \Delta^{-1}(k + 1)$), we need to discuss the following cases:

When $x_{\forall l}^\Pi = x_{\forall l}^\Delta$, the ordinary way is also available because a node in the level $\Delta(x_{\forall l}^\Pi)$ of the partial strategy has only one partial scenario.

When $x_{\forall l}^\Pi \neq x_{\forall l}^\Delta$, we should consider each node $w$ in level $\Delta(x_{\forall l}^\Pi)$. Suppose there are $r$ partial scenarios including the node $w$, and those scenarios are denoted by $(< x_{\Delta^{-1}(1)}, \alpha_{11} >, ..., < x_{\Delta^{-1}(k)}, \alpha_{1k} >), ..., (< x_{\Delta^{-1}(1)}, \alpha_{r1} >, ..., < x_{\Delta^{-1}(k)}, \alpha_{rk} >)$. Suppose $V_k = \{x_{\Delta^{-1}(1)}, ..., x_{\Delta^{-1}(k)}\}$. We divide variables into three sets: $V_d = dif(\Pi, \Delta, x_l)$, $V_s = suc_\forall(\Pi, x_{\forall l}^\Pi)$, and $V_e = V_k - V_d - V_s$. To show that $\bigcap R_{\Delta^{-1}(g)l}(\alpha_{fg})$ ($1 \leq f \leq r, 1 \leq g \leq k$) is not empty, we should prove the proposition that for each pair of $(\alpha_{fg}, \alpha_{hj})$, where $1 \leq f, h \leq r, 1 \leq g, j \leq k$ ($x_{\Delta^{-1}(g)}$ and $x_{\Delta^{-1}(j)}$ may be same), either $R_{\Delta^{-1}(g)l}(\alpha_{fg}) \subseteq R_{\Delta^{-1}(j)l}(\alpha_{hj})$ or $R_{\Delta^{-1}(j)l}(\alpha_{hj}) \subseteq R_{\Delta^{-1}(g)l}(\alpha_{fg})$. Four cases should be considered:

(a) If one of variables $x_{\Delta^{-1}(g)}$ and $x_{\Delta^{-1}(j)}$ is in $V_s$, as $P^\Delta$ is quantified arc consistent, $R_{\Delta^{-1}(g)l}(\alpha_{fg})$ or $R_{\Delta^{-1}(j)l}(\alpha_{hj})$ is $D(x_l)$. So the proposition holds.

(b) If both variables are in $V_e$, as $\Pi$ is semi-compatible with $\Delta$, then $x_{\Delta^{-1}(g)}$ is either in $block(\Pi, x_l)$ or in some block before $block(\Pi, x_l)$. If $g \leq \Delta(x_{\forall l}^\Pi)$, it is clear that $\alpha_{1g} = ... = \alpha_{rg}$; otherwise, $Q(x_{\Delta^{-1}(g)}) = \exists$ and $\Delta(x_{\forall \Delta^{-1}(g)}^\Pi) \leq \Delta(x_{\forall l}^\Pi)$, because the strategy is $(x_{\forall \Delta^{-1}(g)}^\Pi, x_{\Delta^{-1}(g)})$-compatible, we have $\alpha_{1g} = ... = \alpha_{rg}$. So as to $x_{\Delta^{-1}(j)}$. Since $P^\Delta$ satisfies the QBTP, $R_{\Delta^{-1}(g)l}(\alpha_{1g}) \subseteq R_{\Delta^{-1}(j)l}(\alpha_{1j})$ or $R_{\Delta^{-1}(j)l}(\alpha_{1j}) \subseteq R_{\Delta^{-1}(g)l}(\alpha_{1g})$, so the proposition holds.

(c) If one variable is in $V_e$ and the other is in $V_d$, because the $P^\Delta$ satisfies the QBTP and assignments to the variable in $V_e$ are same, it is clear that the proposition holds.

(d) If both variables are in $V_d$, because the triple of variables $(x_{\Delta^{-1}(g)}, x_{\Delta^{-1}(j)}, x_{\Delta^{-1}(l)})$ satisfies the QBAP under $P^\Delta$, then we have $R_{\Delta^{-1}(g)l}(\alpha_{fg}) \subseteq R_{\Delta^{-1}(j)l}(\alpha_{hj})$ or $R_{\Delta^{-1}(j)l}(\alpha_{hj}) \subseteq R_{\Delta^{-1}(g)l}(\alpha_{fg})$.

Hence, $R_{\Delta^{-1}(g)l}(\alpha_{fg})$ ($1 \leq f \leq r, 1 \leq g \leq k$) can be totally ordered, and have a minimal one that is not empty. We then can select a value from the minimal set and create nodes for level $k + 1$ in the subtree of $w$ labeled by the value. Hence, we can construct the level $k + 1$ of the partial strategy, which is $(x_{\forall l}^\Pi, x_l)$-compatible.

Therefore, a solution to $P^\Delta$, which is $(x_{\forall i}^\Pi, x_i)$-compatible for each existentially quantified variable $x_i \in V$, can be constructed. According to Lemma 3, we conclude that $P^\Pi$ is satisfiable. □

Such $P^\Delta$ in Theorem 3 is called a QBTP-adjoint problem of $P^\Pi$.

We provide an instance to show the new hybrid tractable class.

**Example 2.** Consider the problem $P^\Pi = (V, \Pi, D, C)$ with 3 variables, where $Q(x_1) = \exists, Q(x_2) = \forall, Q(x_3) = \exists, \Pi(x_1) = 1, \Pi(x_2) = 2, \Pi(x_3) = 3$, and the constraints are: $R_{12} = \{(\alpha_1, \beta_1), (\alpha_1, \beta_2), (\alpha_2, \beta_1), (\alpha_2, \beta_2)\}$; $R_{13} = \{(\alpha_1, \gamma_1), (\alpha_1, \gamma_2), (\alpha_2, \gamma_2)\}$; $R_{23} = \{(\beta_1, \gamma_1), (\beta_2, \gamma_2)\}$. It can be seen that $P^\Pi$ does not satisfy the QBTP because $(\alpha_2, \beta_1)$ fails to hold the property. Let $\Delta$ be another mapping, where $\Delta(x_1) = 3, \Delta(x_2) = 1, \Delta(x_3) = 2$. $P^\Delta$ satisfies the QBTP and the triple of variables $(x_3, x_3, x_1)$ satisfies the QBAP under $P^\Delta$, so we can construct a solution to $P^\Delta$ (Fig. 3(a)) and then construct a solution to $P^\Pi$ (Fig. 3(b)).

**Theorem 4.** *Given a binary QCSP $P^\Pi = (V, Q, \Pi, D, C)$, there exists a polynomial-time algorithm to find a mapping $\Delta$ such that $P^\Delta = (V, Q, \Delta, D, C)$ is a QBTP-adjoint problem of $P^\Pi$. (or determine that no such mapping exists.)*

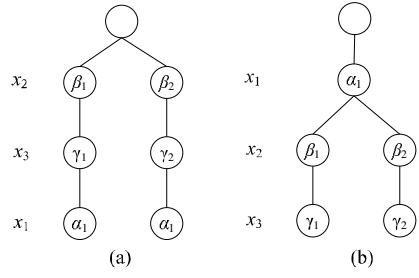

Figure 3: Solutions of Example 2.

*Proof.* We construct a problem whose solution is a suitable variable ordering that is represented by the bijective mapping $\Delta$. First, we add constraints to ensure the semi-compatible property. For each pair of variables $(x_i, x_j)$ such that $Q(x_i) = \forall$, if $\Pi(x_i) < \Pi(x_j)$, we add $\Delta(x_i) < \Delta(x_j)$. Second, to ensure the QBTP, we add the constraint $\Delta(x_k) < max\{\Delta(x_i), \Delta(x_j)\}$ for each triple of variables $(x_i, x_j, x_k)$ if the QBTP fails to hold with the variable ordering $\Delta(x_i) < \Delta(x_j) < \Delta(x_k)$. Third, to ensure the QBAP, we add the constraint $\Delta(x_k) < max\{\Delta(x_i), \Delta(x_j)\}$ for each triple of variables $(x_i, x_j, x_k)$ such that $Q(x_k) = \exists$ and $x_i, x_j \in suc_\exists(\Pi, x_k)$ if the QBAP fails to hold with the variable ordering $\Delta(x_i) \leq \Delta(x_j) < \Delta(x_k)$.

Clearly, the first step requires $O(n^2)$ time. For each triple of variables, checking the QBTP needs $O(d^4)$ time. Because there are $O(n^3)$ triples, the second step needs $O(n^3 d^4)$ time. Similarly, the third step also requires $O(n^3 d^4)$ time. Furthermore, the constraints we added are all max-closed (Dechter 2003), so the problem constructed can be solved in polynomial time. Therefore, $\Delta$ can be found in polynomial time or we can determine that no such mapping exists. $\square$

## The Min-of-max Extendable Class

The min-of-max extendable (MME) class can be regarded as a generalization of the class with the BTP in classical CSPs (Cooper, Jeavons, and Salamon 2010). In this section, we introduce the definition of MME for QCSPs.

**Definition 9.** *Let $P^\Pi = (V, Q, \Pi, D, C)$ be a binary QCSP, where all variable domains are totally ordered. It is min-of-max extendable (QMME for short), if for all triples of variables $(x_i, x_j, x_k)$ such that $\Pi(x_i) < \Pi(x_j) < \Pi(x_k)$, if $(\alpha, \beta) \in R_{ij}$, then assignments $(<x_i, \alpha>, <x_j, \beta>, <x_k, \gamma>)$ are consistent, where $\gamma = min(max(R_{ik}(\alpha)), max(R_{jk}(\beta)))$.*

Following the methods discussed in Definition 8, we give a similar definition on extended MME for QCSPs.

**Definition 10.** *Let $P^\Pi = (V, Q, \Pi, D, C)$ be a binary QCSP, where all variable domains are totally ordered. A triple of variables $(x_i, x_j, x_k)$ is extended QMME under the QCSP $P^\Pi$ if $\Pi(x_i) \leq \Pi(x_j) < \Pi(x_k)$, and for each pair of $(\alpha, \beta)$ such that $\alpha \in D(x_i)$ and $\beta \in D(x_j)$, assignments $(<x_i, \alpha>, <x_j, \beta>, <x_k, \gamma>)$ are consistent, where $\gamma = min(max(R_{ik}(\alpha)), max(R_{jk}(\beta)))$.*

Thus, given the above two definitions, we can identify a new tractable class of QCSPs as follows.

**Theorem 5.** *Let $P^\Pi = (V, Q, \Pi, D, C)$ be a binary quantified arc consistent QCSP whose variable domains are totally ordered. $P^\Pi$ is satisfiable if there exists a mapping $\Delta$ such that*:

1. *$\Pi$ is semi-compatible with $\Delta$;*
2. *$P^\Delta = (V, Q, \Delta, D, C)$ is QMME;*
3. *For each existentially quantified $x_k$, for each pair of variables $(x_i, x_j)$ such that $x_i, x_j \in dif(\Pi, \Delta, x_k)$ and $\Delta(x_i) \leq \Delta(x_j)$, $(x_i, x_j, x_k)$ is extended QMME under $P^\Delta$.*

We omit the detailed proof of Theorem 5, since it directly follows from that of Theorem 3. The BTP and BAP ensure that there exist consistent values for unassigned variables, while MME and extended MME also guarantee that.

## Conclusions

In this paper, we have concentrated on identifying tractable classes of QCSPs by using hybrid techniques, such as the broken-triangle property and min-of-max extendability. First, we have presented the definition of the QBTP on binary QCSPs, which restricts the variable ordering of the QBTP must identical to the ordering in the prefix of the QCSP, and proved that a binary QCSP can be solved in polynomial time if it satisfies the QBTP. We have also discussed the cases that the variable ordering of the QBTP differs from the ordering in the prefix. One simple case that variables can be shifted in their blocks has been discussed, and then a more complex case that existentially quantified variables can be shifted out of their blocks has been studied, which requires the QBAP to ensure the tractability. Moreover, we have extended the tractable results to min-of-max extendable QCSPs.

We regard this work as the first step on hybrid tractable classes of QCSPs. It is noted that there exist other hybrid tractable classes, such as row convex and tree convex constraints, which have some similar tractable properties with the BTP classes while require path consistency to ensure the tractability, so we suggest that more works are needed to discuss the tractable properties of QCSPs with row or tree convex constraints. Furthermore, we also believe that further works are required to investigate more detailed cases so as to identify more tractable classes of QCSPs, as we only give some sufficient conditions to ensure the tractability.

## Acknowledgements

This work was supported by the National Natural Science Foundation of China Granted No. 60803102 and 60973033.